\documentclass[a4paper,french]{rnti}

\usepackage{url}

\usepackage{graphicx}

\usepackage{amsmath, amssymb, amsfonts, nccmath}
\usepackage{array}
\usepackage{caption}
\usepackage{dsfont}
\usepackage{graphicx}
\usepackage[hidelinks]{hyperref}
\usepackage{longtable}  
\usepackage{lipsum}
\usepackage{multirow}
\usepackage{pifont}
\usepackage{soul}
\usepackage{subcaption}
\usepackage[normalem]{ulem}
\usepackage{xcolor}

\titre{Découvrir de nouvelles classes dans des données tabulaires}

\titrecourt{Découvrir de nouvelles classes dans des données tabulaires}

\nomcourt{C. Troisemaine et al.}

\auteur{
    Colin Troisemaine\affil{1}\affilsep\affil{2}
    Joachim Flocon-Cholet\affil{1}
    Stéphane Gosselin\affil{1}\\
    Sandrine Vaton\affil{2}
    Alexandre Reiffers-Masson\affil{2}
    Vincent Lemaire\affil{1}
}

\affiliation{
    \affil{1}Orange Labs, Lannion, France\\
    \affil{2}Département Informatique, IMT Atlantique, Brest, France
 }

\resume{
Dans le domaine du Novel Class Discovery (NCD), le but est de trouver de nouvelles classes dans un ensemble non étiqueté lorsqu'un ensemble étiqueté de classes connues mais différentes est disponible. Bien que le NCD ait récemment attiré l'attention de la communauté scientifique, aucune solution n'a encore été proposée pour les données tabulaires, alors qu'il s'agit d'une représentation très courante des données. Dans cet article, nous proposons une nouvelle méthode pour résoudre ce problème, dans le contexte de données tabulaires contenant des variables hétérogènes. Ce processus est en partie réalisé par une nouvelle méthode de définition de pseudo-étiquettes ainsi que par une mise en \oe{}uvre des découvertes récentes de l'apprentissage multi-tâches pour optimiser une fonction objectif conjointe. Notre méthode démontre que le NCD n'est pas seulement applicable aux images mais aussi aux données tabulaires hétérogènes.
}

\summary{
In Novel Class Discovery (NCD), the goal is to find new classes in an unlabeled set given a labeled set of known but different classes.
While NCD has recently gained attention from the community, no framework has yet been proposed for heterogeneous tabular data, despite being a very common representation of data.
In this paper, we propose TabularNCD, a new method for discovering novel classes in tabular data.
We show a way to extract knowledge from already known classes to guide the discovery process of novel classes in the context of tabular data which contains heterogeneous variables.
A part of this process is done by a new method for defining pseudo labels, and we follow recent findings in Multi-Task Learning to optimize a joint objective function.
Our method demonstrates that NCD is not only applicable to images but also to heterogeneous tabular data.
}

\begin{document}

\setlength{\abovedisplayskip}{3pt}
\setlength{\belowdisplayskip}{3pt}

\section{Introduction}
\label{sec:intro}


Le récent succès des modèles d'apprentissage automatique a été rendu possible en partie par l'utilisation de grandes quantités de données étiquetées. De nombreuses méthodes supposent actuellement qu'une grande partie des données disponibles est étiquetée et que toutes les classes sont connues. Cependant, ces hypothèses ne sont pas toujours vraies en pratique et les chercheurs commencent à envisager des scénarios dans lesquels des données non étiquetées sont disponibles \citep{NodetLBCO21a}.
On peut distinguer dans cet apprentissage dit faiblement supervisé les méthodes qui nécessitent de connaître toutes les classes à l'avance de celles qui sont capables de gérer des classes qui ne sont jamais apparues pendant l'entraînement. 

Récemment, le Novel Class Discovery (NCD) \citep{hsu2018learning} a été proposé pour combler ces lacunes et tente d'identifier de nouvelles classes dans un ensemble de  données non étiquetées en exploitant un autre ensemble étiqueté de classes différentes.
Plusieurs solutions ont été proposées 
dans le contexte de la vision par ordinateur \citep{autonovel2021, han2019learning, zhong2021neighborhood} avec des résultats prometteurs.

Cependant, la recherche dans ce domaine est encore récente et, à notre connaissance, le NCD n'a pas été directement abordé pour les données tabulaires. Bien que les données audio et image suscitent un grand intérêt dans les publications scientifiques récentes, les données tabulaires restent une structure d'information très courante que l'on retrouve dans de nombreux problèmes du monde réel, tels que les systèmes d'information des entreprises. Dans cet article, nous nous concentrerons donc sur le NCD pour les données tabulaires. À la différence des données audio ou image, les données tabulaires sont ``hétérogènes'' et posent certains défis aux modèles d'apprentissage automatique. On peut citer les valeurs manquantes, extrêmes (outliers), erronées ou incohérentes. De plus, le manque de corrélation spatiale entre les attributs rend difficile l'utilisation de techniques basées sur des biais inductifs, telles que les convolutions ou l'augmentation de données. Pour toutes ces raisons, il n'est pas possible de directement transférer les méthodes de NCD conçues pour l'image aux données tabulaires.

\textbf{Notre proposition:} Pour résoudre le problème du NCD dans l'environnement difficile des données tabulaires, nous proposons TabularNCD (pour Tabular Novel Class Discovery). Une représentation latente est d'abord initialisée en tirant parti des avancées du Self-Supervised Learning (SSL) pour les données tabulaires \citep{NEURIPS2020_7d97667a}. Puis, en considérant que le voisinage proche d'une donnée dans l'espace latent est susceptible d'appartenir à la même classe, un partitionnement des données non étiquetées est appris grâce à des mesures de similarité. Le processus de cette deuxième étape est optimisé conjointement avec un classifieur sur les classes connues pour inclure les attributs pertinents des classes déjà découvertes.

\textbf{Travaux connexes.} Les méthodes de NCD se situent à l'intersection de plusieurs domaines dont nous passons certains en revue ici. Le \textit{Transfer Learning} (TL) permet de résoudre un problème plus rapidement ou avec une meilleure généralisation en tirant parti de la connaissance issue d'un problème différent (mais lié). Le NCD peut être considéré comme un problème de TL, cependant comme la majorité des articles de TL nécessitent que toutes les données soient étiquetées, ces méthodes ne peuvent être transférées à notre problème. Un autre domaine proche est le \textit{Semi-Supervised Learning}, où l'on exploite l'ensemble des points d'un jeu de données partiellement étiqueté. Les méthodes de ce domaine supposent que toutes les classes sont connues, ou bien qu'un sous ensemble de chaque classe est représenté dans les données étiquetées, ce qui n'est pas le cas en NCD. Enfin, en \textit{Novelty Detection} (ND), on cherche à prédire si les données font partie des classes connues ou non. En ND, on ne cherche qu'à distinguer le connu de l'inconnu, tandis qu'en NCD on souhaite explorer l'inconnu.

Le présent article est un résumé de l'article publié dans la conférence ICKG 2022 \citep{tabularncdpreprint}.


\section{Découvrir de nouvelles classes dans des données tabulaires}
\label{sec:methode}

Étant donné un ensemble étiqueté $D^l=\{X^l, Y^l\}$ où chaque donnée $x^l$ a une étiquette $y^l \in \{0,1\}^{C^l }$ (représentant l'encodage one-hot des classes $C^l$) et un ensemble sans étiquette $D^u=\{X^u\}$, l'objectif est d'identifier et de prédire les classes de $D^u$. Dans cet article, on suppose que le nombre $C^u$ de classes de $D^u$ est connu et que les classes de $D^l$ et $D^u$ sont disjointes, mais partagent des caractéristiques sémantiques de haut niveau, de sorte que nous pouvons extraire une connaissance de $D^l$ de ce qui constitue une classe pertinente.

La méthode proposée comprend deux étapes principales : une projection des données est d'abord initialisée en pré-entraînant un encodeur $\phi$ sur $D^l \cup D^u$ sans utiliser d'étiquette. Ensuite, une tâche de classification supervisée et une tâche de partitionnement non supervisée sont résolues conjointement sur la représentation précédemment apprise. Chacune de ces deux étapes a sa propre architecture et procédure d'entraînement qui sont décrites ci-dessous.

\textbf{$\bullet$ Initialisation de la représentation.} Cette première étape vise à créer une représentation commune et informative de $D^l$ et de $D^u$, qui n'est pas biaisée envers les données étiquetées.
Ceci est important car la représentation est utilisée à l'étape suivante pour calculer la similarité des paires de données et ainsi déterminer si les exemples doivent appartenir au même cluster ou non.
Une approche plus simple serait d'entraîner un classifieur en utilisant les classes connues. Mais la représentation résultante serait rapidement suradaptée à ces classes et les caractéristiques uniques de $D^u$ seraient perdues.
Pour pré-entraîner l'espace latent à l'aide de toutes les données, étiquetées ou non, nous tirons donc parti du \textit{Self-Supervised Learning} (SSL) et appliquons la méthode \textit{Value Imputation and Mask Estimation} (VIME) \citep{NEURIPS2020_7d97667a}. VIME définit deux tâches pour entraîner l'encodeur $\phi$. À partir d'un vecteur d'entrée $x \in \mathbb{R}^d$ qui a été corrompu, l'objectif est de 1) récupérer les valeurs d'origine et 2) retrouver le masque utilisé pour corrompre $x$.
Le vecteur corrompu $\Tilde{x}$ est généré en remplaçant certaines dimensions par la valeur d'une autre donnée de l'ensemble d'apprentissage choisie aléatoirement.

Suivant l'architecture utilisée dans \citep{NEURIPS2020_7d97667a}, notre encodeur $\phi$ est une combinaison de couches denses avec des fonctions d'activation non linéaires. Pour estimer les valeurs originales et les masques de corruption, deux simples couches denses sont ajoutées après la sortie de l'encodeur. Les autres détails techniques peuvent être trouvés dans l'article de VIME.

\begin{figure*}[tb]
	\centerline{\includegraphics[width=0.85\textwidth]{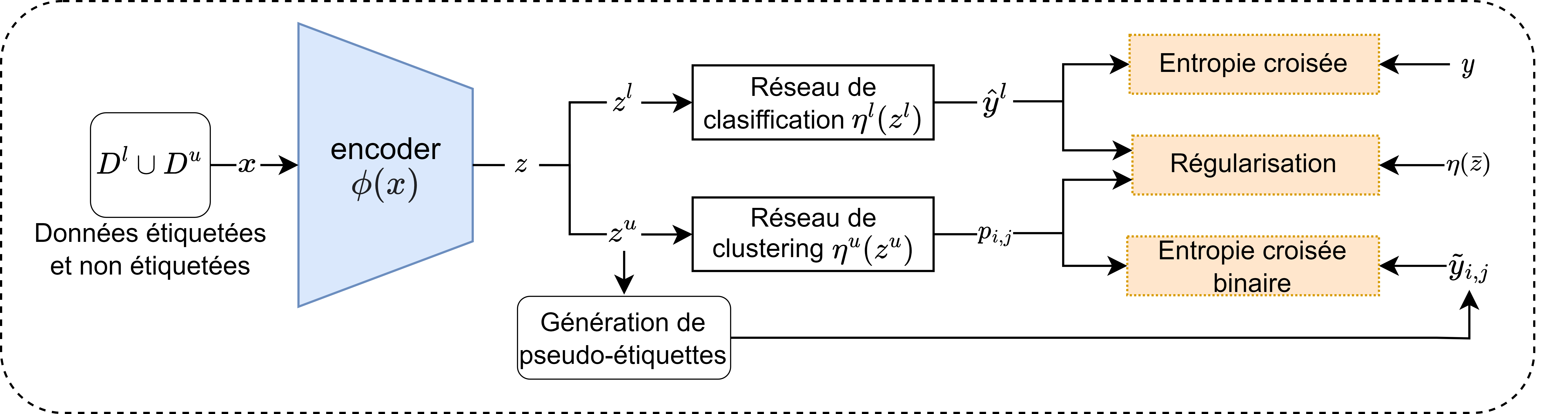}}
    \caption{Architecture de l'apprentissage conjoint.}
    \label{fig:tabularncdmodelstep3}
\end{figure*}

\textbf{$\bullet$ Apprentissage conjoint sur les données étiquetées et non étiquetées} Dans cette étape, deux nouveaux réseaux sont ajoutés à la sortie de l'encodeur précédemment initialisé, chacun résolvant des tâches différentes sur des données différentes (voir Fig.~\ref{fig:tabularncdmodelstep3}). Le premier est un réseau de classification $\eta^l(z) \in \mathbb{R}^{C^l+1}$ entraîné à prédire 1) les $C^l$ classes connues de $D^l$ et 2) une classe unique formée de l'agrégation des données non étiquetées. Le second est un autre réseau de classification entraîné à prédire les $C^u$ nouvelles classes de $D^u$. On l'appellera le réseau de \textit{clustering} $\eta^u(z) \in \mathbb{R}^{C^u}$. Ces deux réseaux partagent le même espace latent et le mettent à jour par rétropropagation, partageant ainsi leurs informations l'un avec l'autre.

Le réseau de classification est optimisé avec l'\textit{entropie croisée} en utilisant les étiquettes $y$ de la vérité terrain : $ l_{class.}= - \sum_{c=1}^{C^l+1} y_{c}\log(\eta_c^l(z)) $, avec $z = \phi(x)$. Son rôle est de guider la représentation pour inclure les caractéristiques des classes connues qui sont pertinentes pour la tâche de classification supervisée.

Pour entraîner le réseau de clustering $\eta^u$ avec des données non étiquetées de manière supervisée, des pseudo-étiquettes $\tilde{y}_{i,j} \in \{0, 1\}$ sont générées pour chaque paire $(x_i, x_j)$ de données non étiquetées dans un mini-batch.
Elles sont définies comme suit : $\tilde{y}_{i,j} = 1$ si $z_i$ et $z_j$ sont similaires, et $\tilde{y}_{i,j} = 0$ sinon.
Pour pouvoir comparer les classes prédites par le réseau de clustering aux $\tilde{y}_{i,j}$, on définit $p_{i,j} = \eta^u(z_i) \cdot \eta^u(z_j)$. Ce score est proche de 1 si le réseau de clustering a prédit la même classe pour $z_i$ et $z_j$, et proche de 0 sinon.
Le réseau de clustering est optimisé avec l'\textit{entropie croisée binaire} :
$ \medmath{
    l_{clust.} = \frac{1}{|Z|-1} \sum\limits_{\substack{j=1 \\ j \neq i}}^{|Z|} \left[ -\tilde{y}_{i,j} \log(p_{i,j}) - (1 - \tilde{y}_{i,j}) \log(1 - p_{i,j}) \right]} $
L'intuition est que des données similaires entre elles dans l'espace latent sont susceptibles d'appartenir à la même classe. Par conséquent, $\eta^u$ créera des clusters de données similaires, guidé par $\eta^l$ avec la connaissance des classes connues.

\textbf{$\bullet$ Définition des pseudo-étiquettes.} Les pseudo-étiquettes $\tilde{y}_{i,j}$ sont définies en fonction de la similarité. Cette idée a été employée dans de nombreux travaux de NCD \citep{autonovel2021, zhong2021neighborhood}, où l'approche la plus courante consiste à définir un seuil $\lambda$ pour la similarité minimale des paires de données à attribuer à la même classe. Cependant, nous avons trouvé\footnote{Expérimentalement, voir le matériel supplémentaire dans \url{https://github.com/ColinTr/TabularNCD}} que définir pour chaque point les $k$ données les plus similaires comme positives était une méthode plus fiable. Ainsi, pour chaque paire $(x_i, x_j)$ dans le batch projeté de données non étiquetées $Z$, les pseudo-étiquettes sont attribuées comme suit :
\begin{equation}
    \tilde{y}_{i,j} = \mathds{1} [ j \in \underset{\substack{r \in \{ 1, ..., |Z| \} \\ r \neq i}}{\text{argtop}_k } \delta(z_i, z_r) ]
    \label{eq:topk_cos_sim}
\end{equation}
où $\delta(z_i, z_r) = \frac{z_i \cdot z_r}{\lVert z_i \rVert \lVert z_r \rVert}$ est la similarité cosinus et $\text{argtop}_k$ est le sous-ensemble d'indices des $k$ plus grands éléments.


\textbf{$\bullet$ Régularisation} Au cours de l'entraînement conjoint des deux réseaux, l'espace latent évolue. Et comme les pseudo-étiquettes sont définies en fonction de la similarité des données dans l'espace latent, cela peut les amener à changer d'une itération à l'autre. Pour limiter ce phénomène, un terme de régularisation est introduit. L'idée est d'inciter le modèle à prédire la même classe pour une donnée $x$ et sa contrepartie perturbée $\bar{x}$.
Ici, nous utilisons SMOTE-NC \citep{SMOTE} pour créer les nouvelles données perturbés.
L'\textit{erreur quadratique moyenne} (MSE) est utilisée comme terme de régularisation pour les réseaux de classification et de clustering. Pour le réseau de clustering, il s'écrit : $ l_{reg.} = \frac{1}{C^u} \sum_{c=1}^{C^u} (\eta_c(z) - \eta_c(\bar{z}))^2 $, où $\bar{z}=\phi(\bar{x})$ est la projection de $x$ perturbé avec SMOTE-NC. Pour le réseau de classification, la moyenne est réalisée sur $C^l+1$.

\textbf{$\bullet$ Fonction de perte complète} Plutôt que de définir la fonction de perte comme une somme pondérée de tous les objectifs, il a été choisi de définir une fonction de perte et un optimiseur pour chacun des réseaux. Ainsi, la perte du réseau de classification est :
$\mathcal{L}_{classification} = w_1 l_{classif.}+ (1- w_1) l_{reg.}$
Et la perte du réseau de clustering est : $\mathcal{L}_{clustering} = w_2 l_{clust.} + (1 - w_2) l_{reg.}$, où $w_1$ et $w_2$ sont des hyper-paramètres permettant d'équilibrer le poids des termes de régularisation. Les deux réseaux sont entraînés de manière alternée : Pour chaque mini-batch, le réseau de classification et l'encodeur sont d'abord mis à jour par rétropropagation avec $\mathcal{L}_{classification}$. Ensuite, les données non étiquetées du même mini-batch sont utilisées pour calculer $\mathcal{L}_{clustering}$, qui est rétropropagé pour mettre à jour le réseau de classification et l'encodeur une fois de plus.

\section{Expériences}

\subsection{Jeux de données et détails expérimentaux}
\label{sec:experimental_details}

\textbf{$\bullet$ Jeux de données.} Pour évaluer les performances de la méthode proposée ici, six jeux de données tabulaires de classification ont été choisis (voir Table~\ref{table:datasets_description}), ainsi que MNIST, dont les images ont été aplaties pour former des vecteurs de $28 \times 28 = 784$ attributs. Si les données d'entraînement et de test ne sont pas déjà séparées, 70 $\%$ sont conservés pour l'entraînement, tandis que les 30 $\%$ restants sont utilisés en test.
Suivant la même procédure que dans les articles de Novel Class Discovery \citep{autonovel2021, zhong2021neighborhood}, nous cachons les étiquettes d'environ 50\% des classes pour créer les classes \textit{inconnues}. Nous évaluons ensuite la capacité des méthodes comparées à retrouver ces classes. Les partitions des 7 jeux de données sont présentées dans la table \ref{table:datasets_description}.

\begin{table}[tb]
    \fontsize{7}{8}\selectfont
    \caption{Informations statistiques des jeux de données sélectionnés.}
    \begin{center}
    \setlength{\tabcolsep}{2pt}
        \begin{tabular}{l|c|c|c|c|c|c}
            \hline
            \multirow{2}{*}{Nom} & Attri- & \# classes & \# train & \# train & \# test & \# test  \\
                                 & buts & $C^l$ / $C^u$ & étiqueté & non étiqueté & étiqueté & non étiqueté \\
            \hline
            MNIST              & 784   & 5 / 5  & 30,596 & 29,404 & 5,139  & 4,861  \\
            Forest Cover type  & 54    & 4 / 3  & 6,480  & 4,860  & 36,568 & 13,432 \\
            Letter recognition & 16    & 19 / 7 & 10,229 & 3,770  & 4,296  & 1,704  \\
            Human activity     & 562   & 3 / 3  & 3,733  & 3,619  & 1,494  & 1,453  \\
            Satimage           & 36    & 3 / 3  & 2,525  & 1,976  & 1,042  & 887    \\
            Pendigits          & 16    & 5 / 5  & 3,777  & 3,717  & 1,764  & 1,734  \\
            1990 US Census     & 67    & 12 / 6 & 50,000 & 50,000 & 31,343 & 18,657 \\
            \hline
        \end{tabular}
        \label{table:datasets_description}
    \end{center}
\end{table}

\textbf{$\bullet$ Métriques d'évaluation.} Pour évaluer la performance des méthodes comparées, nous utilisons la \textit{clustering accuracy} (ACC) et la \textit{balanced accuracy} (BACC), que l'on calcule après affectation optimale des étiquettes (via l'algorithme hongrois \citep{Kuhn55hungarian}). L'information mutuelle normalisée (NMI) et l'indice Rand ajusté (ARI) sont aussi calculés afin de mesurer la correspondance et la similarité entre deux partitionnements.
Les métriques rapportées dans la section suivante sont toutes calculées sur les jeux de test non étiquetés.

\textbf{$\bullet$ Méthodes concurrentes.} À notre connaissance, il n'existe aucune autre méthode qui résout le problème particulier de Novel Class Discovery pour les jeux de données tabulaires. Néanmoins, nous pouvons nous comparer aux méthodes de clustering non supervisées. Cela nous permettra également de montrer que notre méthode fournit un moyen efficace d'incorporer des connaissances issues de classes connues. Nous choisissons l'algorithme $k$-means pour sa simplicité et sa popularité, ainsi que la méthode de Clustering Spectral \citep{Luxburg07} pour ses bons résultats connus. Nous définissons $k = C^u$ (vérité terrain) pour les deux méthodes de clustering, car il était déjà supposé que $C^u$ soit connu dans la méthode proposée.

Nous définissons également une méthode de base simple qui utilise les classes connues pour partitionner les données non étiquetées : (i) un réseau de neurones de classification est d'abord entraîné sur les classes connues de $D^l$ ; (ii) puis l'avant-dernière couche du classifieur est utilisée comme projection pour les données de $D^u$. Dans cette projection, un $k$-means est appliqué pour attribuer des étiquettes aux données de test. Par rapport aux approches non supervisées, cette technique a l'avantage d'intégrer les caractéristiques des classes connues dans un espace latent de dimensionnalité réduite. Cependant, on s'attend à ce que cette méthode fonctionne de manière sous-optimale car les caractéristiques spécifiques aux données non étiquetées pourraient être perdues dans la dernière couche cachée après l'entraînement.

\textbf{$\bullet$ Détails d'implémentation.} Les hyper-paramètres sont optimisés sur un ensemble de validation représentant $20\%$ de l'ensemble d'entraînement. Lors de l'étape de \textit{Self Supervised Learning}, nous utilisons les valeurs de l'article original de VIME \citep{NEURIPS2020_7d97667a}. Dans l'étape d'apprentissage conjoint, nous utilisons une taille de batch de 512. Les hyper-paramètres sont ici le taux d'apprentissage des deux optimiseurs, ainsi que leurs paramètres $w_1$ et $w_2$ d'équilibre.
Le nombre $top$ $k$ de paires positives par instance et les $k$ $voisins$ considérés dans la méthode d'augmentation de données sont également optimisés.
Nous avons implémenté notre méthode sous Python avec la librairie PyTorch.
Le code est accessible à l'adresse \url{https://github.com/ColinTr/TabularNCD} et les autres détails d'implémentation sont disponibles dans \citep{tabularncdpreprint}.

\subsection{Résultats}

\begin{table}[tb]
    \caption{Performances de TabularNCD sur les classes inconnues.}
    \begin{center}
        \fontsize{7}{7}\selectfont
        \setlength{\tabcolsep}{3pt}
        \begin{tabular}{l | l c c c c}
            \hline
            Jeu de données & Méthode & BACC (\%) & ACC (\%) & NMI & ARI \\
            \hline
            \multirow{4}{*}{MNIST}      & Baseline         & 57.7$\pm$4.7          & 57.6$\pm$4.5          & 0.37$\pm$0.2           & 0.31$\pm$0.3           \\
                                        & Spect. clust     & -                     & -                     & -                      & -                      \\
                                        & \textit{k}-means & 60.1$\pm$0.0          & 61.1$\pm$0.0          & 0.48$\pm$0.0           & 0.38$\pm$0.0           \\
                                        & TabularNCD       & \textbf{91.5$\pm$4.1} & \textbf{91.4$\pm$4.2} & \textbf{0.82$\pm$0.06} & \textbf{0.81$\pm$0.04} \\
            \hline
            \multirow{4}{*}{Forest}     & Baseline         & 55.6$\pm$2.0          & 68.5$\pm$1.4          & 0.27$\pm$0.02          & 0.15$\pm$0.01          \\
                                        & Spect. clust     & 32.1$\pm$1.4          & 85.8$\pm$4.0          & 0.01$\pm$0.01          & 0.09$\pm$0.01          \\
                                        & \textit{k}-means & 32.9$\pm$0.0          & 62.0$\pm$0.0          & 0.04$\pm$0.00          & 0.05$\pm$0.00          \\
                                        & TabularNCD       & \textbf{66.8$\pm$0.6} & \textbf{92.2$\pm$0.2} & \textbf{0.37$\pm$0.09} & \textbf{0.56$\pm$0.09} \\
            \hline
            \multirow{4}{*}{Letter}     & Baseline         & 55.7$\pm$3.6          & 55.9$\pm$3.6          & 0.49$\pm$0.04          & 0.33$\pm$0.04          \\
                                        & Spect. clust     & 45.3$\pm$4.0          & 45.3$\pm$4.0          & 0.48$\pm$0.03          & 0.18$\pm$0.03          \\
                                        & \textit{k}-means & 50.2$\pm$0.6          & 49.9$\pm$0.6          & 0.40$\pm$0.01          & 0.28$\pm$0.01          \\
                                        & TabularNCD       & \textbf{71.8$\pm$4.5} & \textbf{71.8$\pm$4.5} & \textbf{0.60$\pm$0.04} & \textbf{0.54$\pm$0.04} \\
            \hline
            \multirow{4}{*}{Human}      & Baseline         & 80.0$\pm$0.5          & 78.0$\pm$0.6          & 0.64$\pm$0.01          & 0.62$\pm$0.01          \\
                                        & Spect. clust     & 70.2$\pm$0.0          & 69.4$\pm$0.0          & 0.72$\pm$0.00          & 0.60$\pm$0.00          \\
                                        & \textit{k}-means & 75.3$\pm$0.0          & 77.0$\pm$0.0          & 0.62$\pm$0.00          & 0.59$\pm$0.00          \\
                                        & TabularNCD       & \textbf{98.9$\pm$0.2} & \textbf{99.0$\pm$0.2} & \textbf{0.95$\pm$0.01} & \textbf{0.97$\pm$0.01} \\
            \hline
            \multirow{4}{*}{Satimage}   & Baseline         & 53.8$\pm$3.4          & 53.9$\pm$4.2          & 0.25$\pm$0.03          & 0.22$\pm$0.03          \\
                                        & Spect. clust     & 82.2$\pm$0.1          & 77.8$\pm$0.1          & 0.51$\pm$0.00          & 0.46$\pm$0.00          \\
                                        & \textit{k}-means & 73.7$\pm$0.3          & 69.2$\pm$0.2          & 0.30$\pm$0.00          & 0.28$\pm$0.00          \\
                                        & TabularNCD       & \textbf{90.8$\pm$4.0} & \textbf{91.4$\pm$5.0} & \textbf{0.71$\pm$0.11} & \textbf{0.79$\pm$0.07} \\
            \hline
            \multirow{4}{*}{Pendigits}  & Baseline         & 72.8$\pm$5.5          & 72.8$\pm$5.4          & 0.62$\pm$0.06          & 0.54$\pm$0.07          \\
                                        & Spect. clust     & {84.0$\pm$0.0}        & {84.0$\pm$0.0}        & \textbf{0.78$\pm$0.00} & 0.67$\pm$0.00          \\
                                        & \textit{k}-means & 82.5$\pm$0.0          & 82.5$\pm$0.0          & 0.72$\pm$0.00          & 0.63$\pm$0.00          \\
                                        & TabularNCD       & \textbf{85.5$\pm$0.7} & \textbf{85.6$\pm$0.8} & 0.76$\pm$0.02          & \textbf{0.71$\pm$0.02} \\
            \hline
            \multirow{4}{*}{Census}     & Baseline         & 53.0$\pm$3.5          & \textbf{55.0$\pm$6.5} & 0.49$\pm$0.02          & 0.30$\pm$0.03          \\
                                        & Spect. clust     & 23.6$\pm$3.3          & 51.3$\pm$5.5          & 0.24$\pm$0.11          & 0.18$\pm$0.09          \\
                                        & \textit{k}-means & 38.5$\pm$2.6          & 49.8$\pm$3.6          & 0.41$\pm$0.05          & 0.28$\pm$0.03          \\
                                        & TabularNCD       & \textbf{61.9$\pm$0.6} & 50.1$\pm$0.9          & 0.48$\pm$0.01          & 0.30$\pm$0.00          \\
            \hline
        \end{tabular}
        \label{table:tabularncd_results}
    \end{center}
    \footnotesize{L'écart type est calculé sur 10 exécutions. Les 2 méthodes de clustering non supervisées (Spect. clust et $k$-means) ne sont entraînées que sur données de test appartenant aux classes inconnues.
    Les valeurs pour le partitionnement spectral de MNIST sont manquantes car l'exécution ne s'est pas terminée en moins d'une heure.}
\end{table}

\textbf{Comparaison avec les méthodes concurrentes.} Dans le tableau \ref{table:tabularncd_results}, nous rapportons les performances des 4 méthodes concurrentes sur les 7 jeux de données pour la tâche de clustering. Les résultats montrent que TabularNCD atteint des performances supérieures à la méthode de base et aux deux méthodes de clustering non supervisées, sur tous les jeux de données considérés et pour toutes les métriques. Les améliorations de la précision par rapport aux méthodes concurrentes varient entre 1,6\% et 21,0\%. Cela prouve que même pour les données tabulaires, des connaissances utiles peuvent être extraites de classes déjà découvertes.
Alors que la méthode de base ($k$-means sur une projection apprise sur les données étiquetées, voir Sec.~\ref{sec:experimental_details}) améliore les performances de $k$-means pour certains jeux de données (notamment \textit{Census} et \textit{Forest}), elle est encore inférieure à notre méthode en général. Elle obtient un score inférieur au simple $k$-means sur 3 jeux de données, ce qui signifie que dans certains cas, les caractéristiques extraites des classes connues sont insuffisantes pour discriminer de nouvelles données. C'est le cas pour le jeu de données \textit{Satimage}, où les performances de la méthode de base sont bien inférieures à celles du simple $k$-means sur les données d'origine. Pour s'assurer que les classes connues et inconnues partagent des caractéristiques communes, nous référons le lecteur à l'article \citep{li2022a} qui tente de mesurer la similarité sémantique des classes de $D^l$ et $D^u$.

\begin{figure*}[tb]
	\begin{center}
	    \begin{subfigure}{0.24\textwidth}
            \captionsetup{font={footnotesize}}
            \centering
            \includegraphics[width=\textwidth]{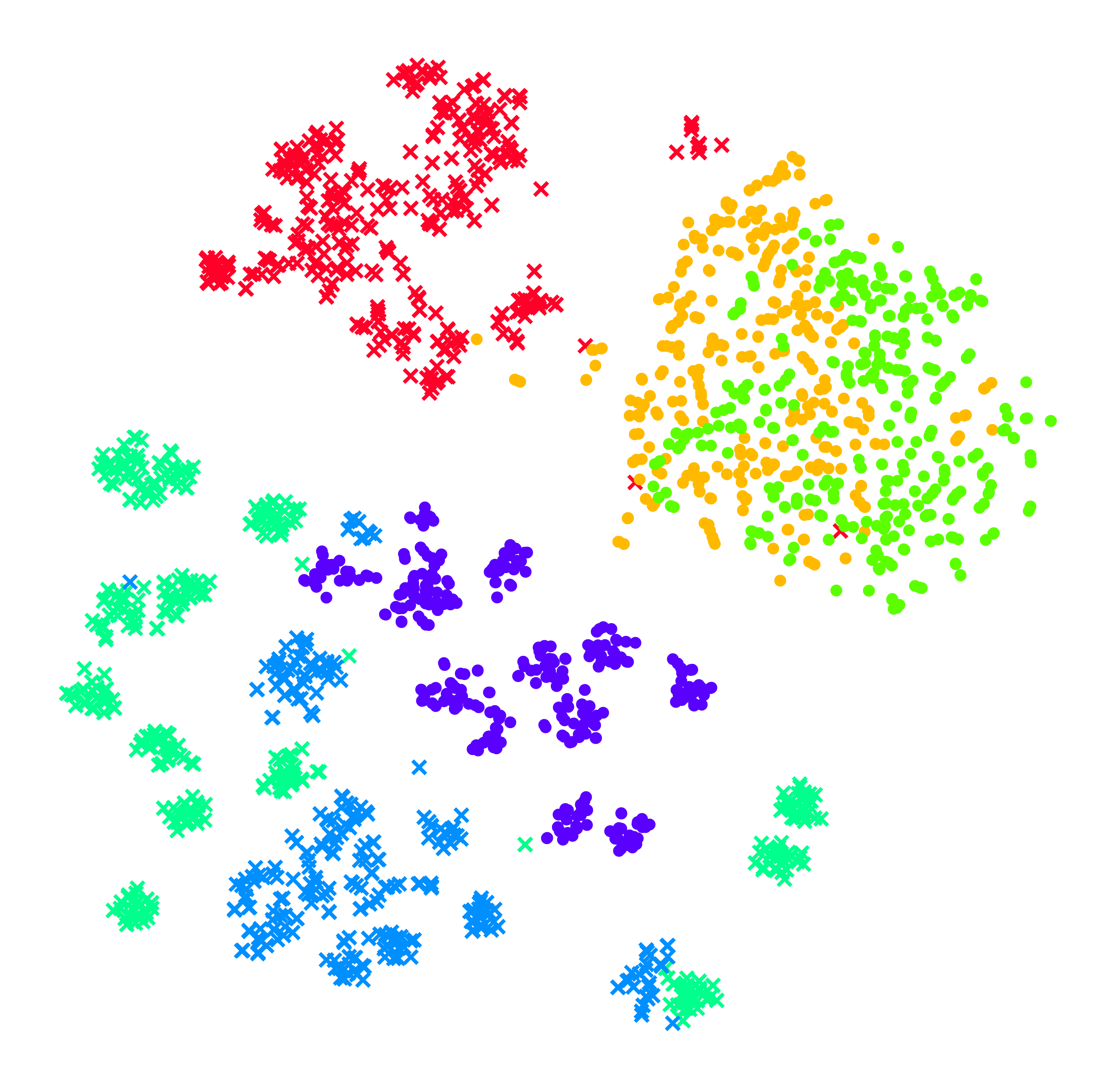}
            \caption{Données originales.}
            \label{fig:latent_init}
        \end{subfigure}
        \hfill
	    \begin{subfigure}{0.24\textwidth}
            \captionsetup{font={footnotesize}}
            \centering
            \includegraphics[width=\textwidth]{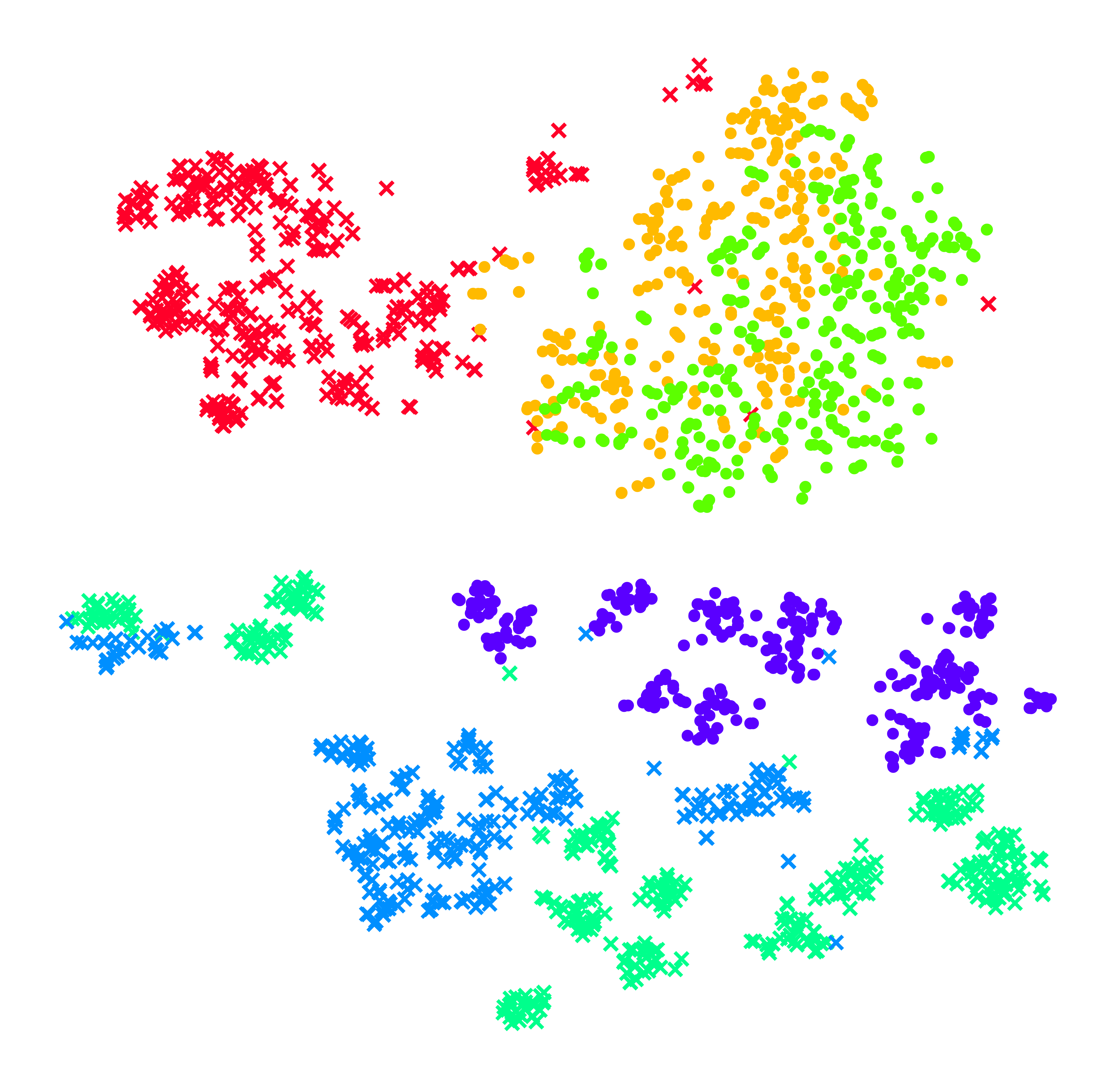}
            \caption{Après SSL.}
            \label{fig:latent_ssl}
        \end{subfigure}
        \hfill
	    \begin{subfigure}{0.24\textwidth}
            \captionsetup{font={footnotesize}}
            \centering
            \includegraphics[width=\textwidth]{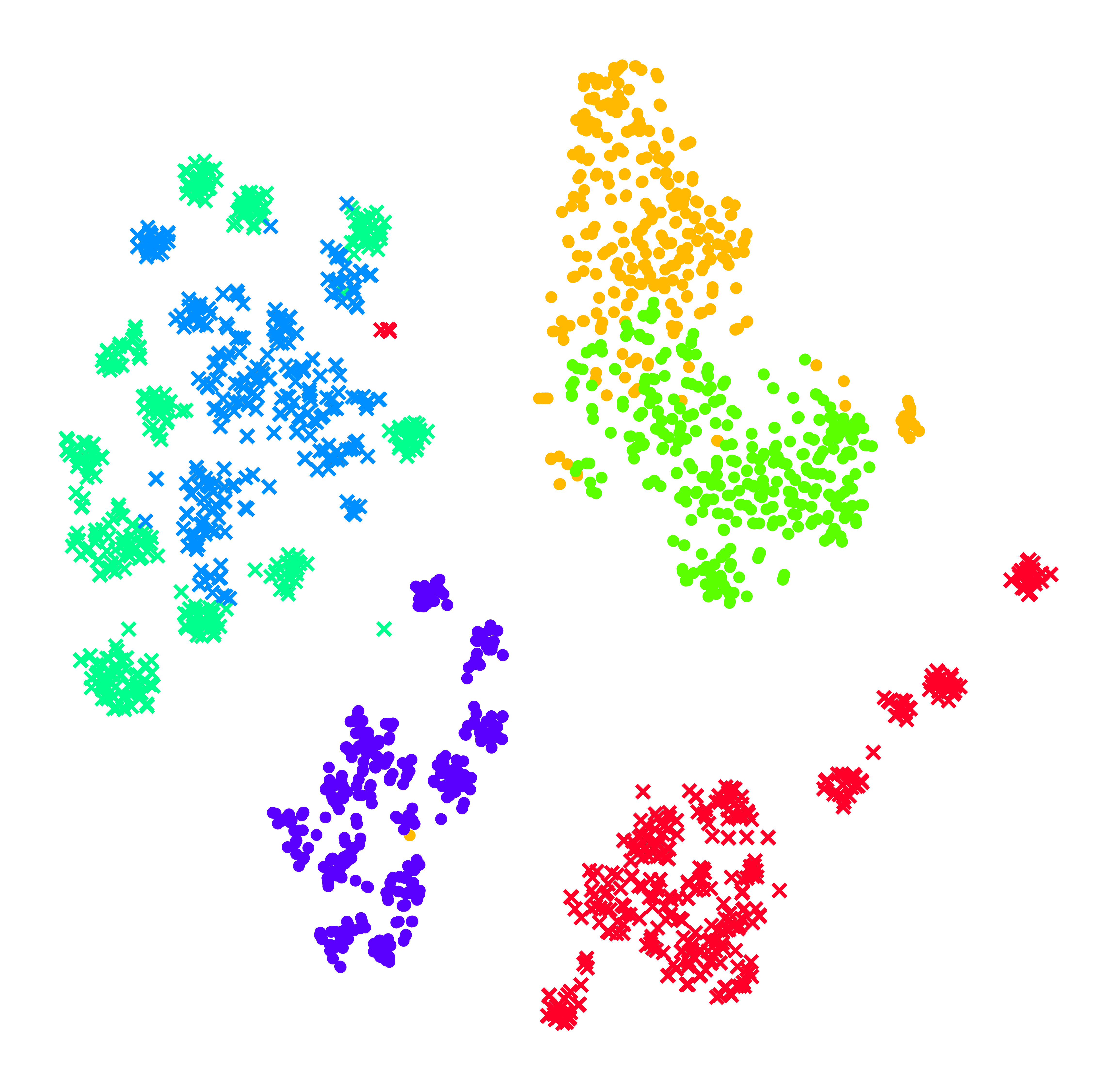}
            \caption{Étape 15 de l'entraînement joint.}
            \label{fig:latent_15}
        \end{subfigure}
        \hfill
	    \begin{subfigure}{0.24\textwidth}
            \captionsetup{font={footnotesize}}
            \centering
            \includegraphics[width=\textwidth]{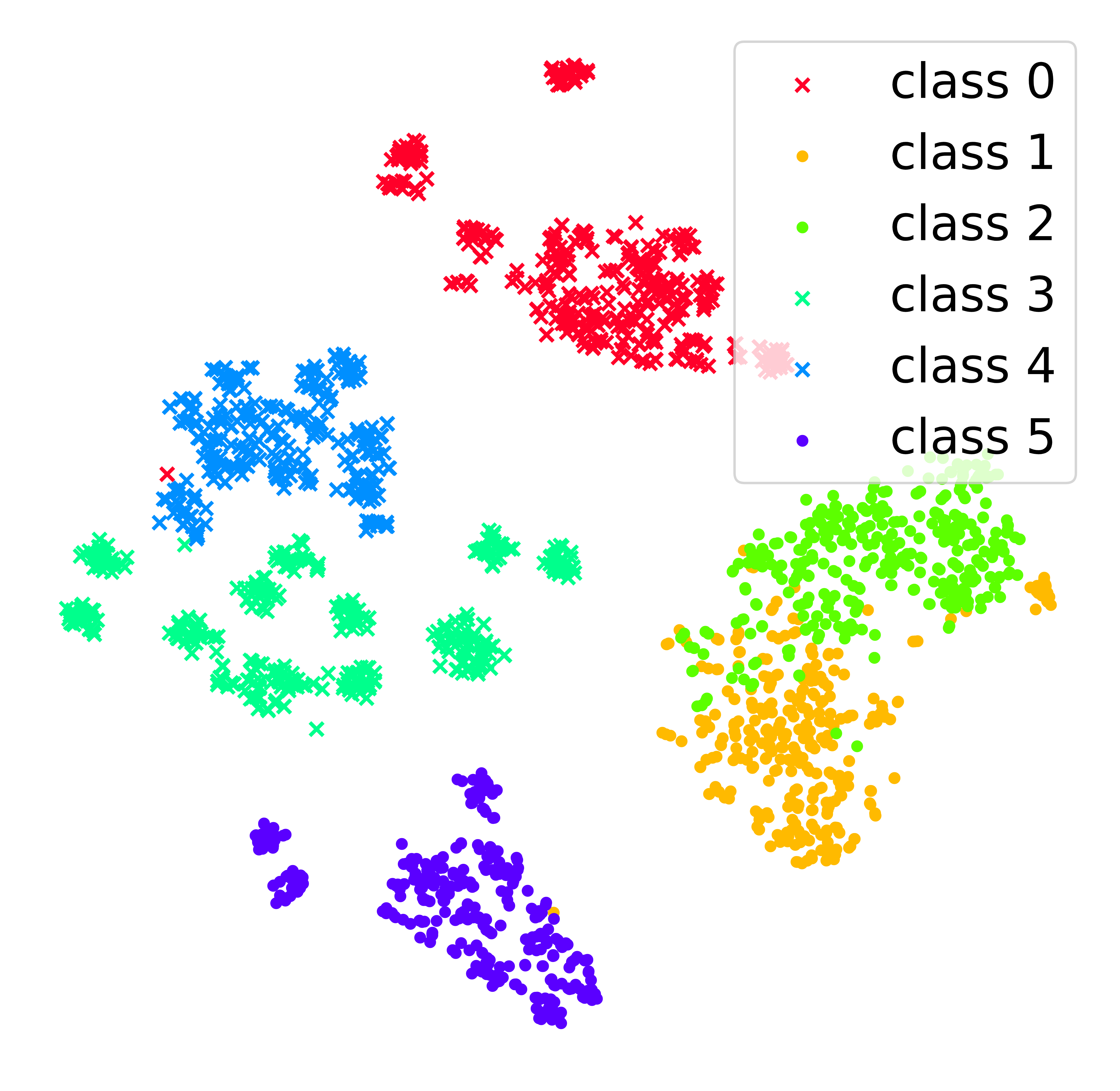}
            \caption{Étape 30 de l'entraînement joint.}
            \label{fig:latent_30}
        \end{subfigure}
        \caption{Évolution du t-SNE lors de l'entraînement conjoint du modèle sur le jeu de données \textit{Human Activity Recognition}.}
        \label{fig:latent_space}
	\end{center}
\end{figure*}

\textbf{Visualisation.} En plus des résultats quantitatifs, nous réalisons une analyse qualitative montrant l'espace des caractéristiques qui est appris. Dans la Fig.~\ref{fig:latent_space}, nous visualisons l'évolution de la représentation de toutes les classes au cours de l'entraînement de la méthode proposée. Fig~\ref{fig:latent_init} correspond aux données d'origine, tandis que les figures suivantes sont les données dans l'espace latent.
Après l'étape de SSL (Fig~\ref{fig:latent_ssl}), les classes se chevauchent toujours, mais cette étape présente tout de même l'avantage d'avoir initialisé l'encodeur à une meilleure représentation que l'aléatoire. Cela signifie qu'au début de l'entraînement conjoint, les pseudo étiquettes définies à l'aide de l'équation \eqref{eq:topk_cos_sim} seront plus exactes. Au cours de l'entraînement conjoint dans les figures \ref{fig:latent_15} et \ref{fig:latent_30}, les classes sont de plus en plus séparées, ce qui indique que la méthode proposée est capable de découvrir avec succès de nouvelles classes en exploitant les connaissances des classes connues. Cette figure montre que notre modèle produit des représentations où les échantillons de la même classe sont étroitement regroupés.

\section{Conclusions et travaux futurs}
Dans cet article, nous avons proposé une première solution au problème de la découverte de nouvelles classes dans l'environnement difficile des données tabulaires. Nous avons démontré l'efficacité de l'approche que nous proposons, TabularNCD, par le biais d'une analyse poussée sur 7 jeux de données contre des méthodes de clustering non supervisées. Les meilleures performances de notre méthode ont montré qu'il est possible d'extraire des connaissances de classes déjà découvertes pour guider le processus de découverte de nouvelles classes, ce qui démontre que NCD n'est pas seulement applicable aux images mais aussi aux données tabulaires. Enfin, la méthode originale de définition de pseudo-étiquettes proposée ici s'est avérée fiable même en présence de classes déséquilibrées. Les avancées récentes de l'apprentissage profond sur les données tabulaires méritent d'être étudiées dans des travaux futurs. En particulier, nous explorerons les réseaux adversariaux génératifs et les encodeurs automatiques variationnels comme substituts à l'encodeur simple du modèle actuel, qui est un élément central de notre méthode. En outre, l'hypothèse selon laquelle le nombre de nouvelles classes est connu est une limitation de notre méthode, et sera certainement une voie future de notre travail.

\bibliographystyle{rnti}
\bibliography{biblio}

\Fr

\end{document}